# S&P 500 Stock's Movement Prediction using CNN

Rahul Gupta
Stanford University
353 Serra Mall, Stanford, CA 94305
rahulgup@stanford.edu

**Abstract**

*This paper is about predicting the movement of stock consist of S&P 500 index. Historically there are many approaches have been tried using various methods to predict the stock movement and being used in the market currently for algorithm trading and alpha generating systems using traditional mathematical approaches [1, 2]. The success of artificial neural network recently created a lot of interest and paved the way to enable prediction using cutting-edge research in the machine learning and deep learning. Some of these papers have done a great job in implementing and explaining benefits of these new technologies. Although most these papers do not go into the complexity of the financial data and mostly utilize single dimension data, still most of these papers were successful in creating the ground for future research in this comparatively new phenomenon.*

*In this paper, I am trying to use multivariate raw data including stock split/dividend events (as-is) present in real-world market data instead of engineered financial data. Convolution Neural Network (CNN), the best-known tool so far for image classification, is used on the multi-dimensional stock numbers taken from the market mimicking them as a vector of historical data matrices (read images) and the model achieves promising results. The predictions can be made stock by stock, i.e., a single stock, sector-wise or for the portfolio of stocks.*

## 1. Introduction

The Standard & Poor's 500 Index or S&P 500 as it called is a weighted index of the 500 largest U.S. publicly traded companies market capitalization. S&P 500 is one of the furthermost commonly quoted American indexes because it is representative of the largest U.S. public corporations and it focuses on the large-cap sector of the U.S. market. [3]

Financial institutions, banks, dealer broker firms, hedge funds and individual rely on S&P 500 stock's movement to invest, liquidate their investments or hedge their business risks. S&P 500 represent top 500 large cap companies of the United States and is watched by companies and individual not just in the US but around the globe as one of the key financial indicators for market movements. Though the same proposed model can be trained for all the stocks listed in S&P 500, for illustration purposes, in this paper the proposed model will use historical prices of few of S&P 500 stocks to train and test the model.

1.1. Background:

In the recent years, deep learning using Artificial Neural Networks (ANN) have been used in a lot of different areas such, to name few such as automotive, diagnostic, surgery, etc. ANN application list is ever increasing. After the ImageNet challenges where Convolutional Neural Network (CNN) based model have made significant progress in predicting images (in fact better than human) and overdo other available approaches used till date, there have been several applications tried using CNN. Another Deep Learning approach, LSTM (stands for Long-Short-Term-Memory), a prominent variation of Recurrent Neural Network or RNN, has been used in predicting time-series data and has provided surprising results. A recent study shows that LSTM outperforms traditional-based algorithms such as ARIMA model [4].

The application of deep learning approaches mainly LSTM and CNN to finance has received a great deal of attention from both investors and academia. LSTMs are challenging to train, mainly because of the computationally expensive nature of time-based gradient descent, the size of the networks, and the amount of data over which they must be trained. Further complicating, since LSTMs are stateful, many problems require "online" training, meaning that they cannot be trained all at once by highly optimized, vectorized calculations in a batch with error computation and gradient descent over groups [5]. However, CNNs with their ability to learn useful spatial features from the input data have revolutionized computer vision. The network topology of CNNs exploits the spatial relationship among the pixels in an image, and this is one of the reasons for their success [6].

Stock market prediction is usually considered as one of the most challenging issues among time series predictions [7] due to its noise, and volatile features make CNN natural choice over LSTM. In addition to the data noise, due to its nature of convolving CNN intuitively would be better of handling prediction of a stock dividend and split events.

The input to our algorithm is a stock's raw historical time-series numbers (OPEN, HIGH, LOW, CLOSE, VOLUME, ADJ_OPEN, ADJ_HIGH, ADJ_LOW, ADJ_CLOSE, ADJ_VOLUME) uploaded in CSV format for all current stock consists in S&P 500. We then use a CNN approach to deep learning to predict BUY/SELL based on the absolute return.

## 2. Related Works

Persio and Honchar[8] laid the framework for comparing different neural network models for stock prediction, concerning the forecast of their trend movements up or down, in their paper Artificial Neural Networks architectures for stock price prediction: comparisons and applications. Like our data set, the S&P 500 historical time series, predicting a trend based on data from the past days, and proposing a novel state-of-art approach based on a combination of wavelets and CNN, which outperforms the basic neural networks ones. Their paper demonstrates that neural networks can predict financial time series movements even trained only on plain time series data.

In their paper, Financial Time Series Prediction using Deep Learning, Ariel Navon, and Yosi Kellery [9] presented a data-driven end-to-end Deep Learning approach for time series prediction, applied to financial time series. In their paper, a Deep Learning scheme is derived to predict the temporal trends of stocks and ETFs in NYSE or NASDAQ. Their approach is based on an artificial neural network to predict the temporal patterns of stocks and ETFs that are applied to raw financial data inputs which very similar to the input data being used in this paper as it used raw market data of indices/ticker from S&P 500.

Convolutional neural networks (CNNs) with their ability to learn useful spatial features have revolutionized computer vision. In Predicting Time Series with Space-Time Convolutional and Recurrent Neural Networks, Grob et al. applied CNNs to predict electricity prices at the European Power Exchange discovering that the CNN's were able to find features with better predictive power than those engineered by an expert. Similar to the way proposed in this paper, in their article, Grob et al. [10] demonstrated how multivariate time series could be interpreted as space-time pictures, thus expanding the applicability of the tricks-of-the-trade for CNNs to this critical domain. Their multidimensional time series approach leveraged the high correlation between intraday electricity contracts. Also, they tested two novel models - Space-Time Convolutional Neural Network (ST-CNN) and Space-Time Convolutional and Recurrent Neural Network (STaR).

In the paper titled as Conditional Time Series Forecasting with Convolutional Neural Networks [11], Borovykh et al. have presented a method for conditional time series forecasting based on the deep convolutional WaveNet architecture. The network used in their model contains stacks of dilated convolutions that allow it to access a wide-ranging history when forecasting; several convolutional filters are applied in parallel to separate time series and allow for the fast processing of data and the exploitation of the correlation structure between the multidimensional time series. The performance of the CNN is analyzed on various multivariate time series and compared to that of the well-known autoregressive model and a long-short-term memory network.

Pinheiro and Dras [12] explores recurrent neural networks with character-level language model pre-training for both intraday and daily stock market forecasting and predicting directional changes in the S&P 500 index, both for individual stocks/tickers and the overall index in their paper Stock Market Prediction with Deep Learning: A Character-based Neural Language Model for Event-based Trading.

Selvin et al. [13] have proposed method that does fit the data to a specific model, instead identify the underlying dynamics existing in the data using deep learning architectures. In their work, they used three different deep learning architectures for the price prediction of NSE listed companies and compares their performance by applying a sliding window approach for predicting future values. The model proposed in this paper also uses the sliding window approach.

In a paper submitted by Siripurapu A. [14] where using convolutional networks to predict movements in stock prices; however, by converting the time series of past price fluctuations to an actual image was a smart approach but requires additional work that is an overhead make it less usable.

Bi´nkowski et al. [15] proposed model they called Significance-Offset Convolutional Neural Network, a deep convolutional network architecture for regression of multivariate asynchronous time series. The model is inspired by standard autoregressive (AR) models and gating mechanisms used in recurrent neural networks.

Sezer and Ozbayoglu [16] proposed a novel algorithmic trading model CNN-TA using a 2-D Convolutional Neural Network based on image processing properties. They converted financial time series into 2-D images and then labeled as Buy, Sell or Hold depending on the hills and valleys of the original time series.

## 3. Dataset and Features

The dataset we are using in this paper and model is raw data and not the engineered financial data. An open-high-low-close-volume (or OHLCV in short) with its adjusted counterpart is used for training and testing this model. The OHLCV is used to illustrate movements in the price of a financial instrument over time along with volume traded over one unit of time, e.g., one day or one hour. For this paper, we are using daily.

The image in figure 1(a) is an example of OHLC candle price chart and 1(b) a zoom-in illustration of a single candle and how to decode it.

An adjusted OHLC price is a stock's price on an any given day of trading that has been amended to include any dividend distributions and corporate actions such as dividend/split that occurred at any time before the next day's open.

OHLCV and adjusted OHLCV are most critical raw numbers that a stock/equity has prices comprises along with few others such as market capitalization, earning, etc.

For the ohlc price, it also needs to be split-unadjusted. An example of unsplit-adjusted price could be found in figure 2 that is part of the dataset used in training.

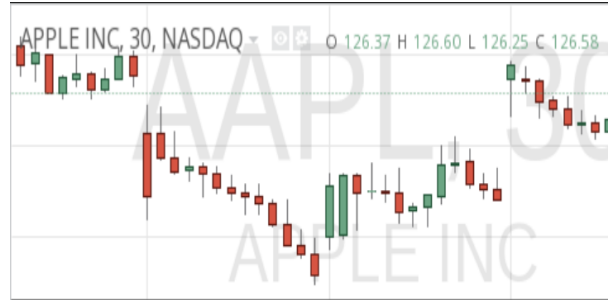

Figure 1 (a): Example of OHLC candle price chart. Image source: tradingview.com/wiki

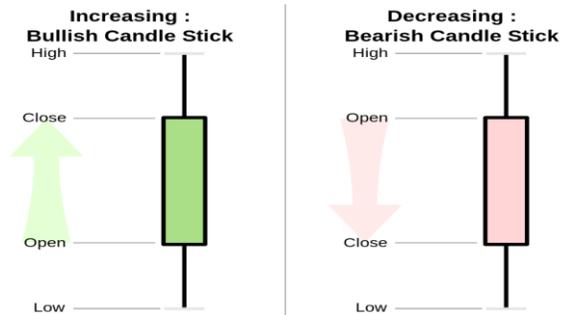

Figure 1 (b): Decoding candle chart. Image source: (b) wikipedia

Stock dividends are like cash dividends and effects company's stock price; however, instead of cash, a company pays out stock [17]. As a result, a company's shares outstanding will increase, and the company's stock price will decrease. On the other hand, stock splits occur when a company perceives that its stock price may be too high and want to keep their stock price within an optimal trading range. Stock splits are usually done to increase the liquidity of the stock and to make it more affordable for investors to buy regular lots.

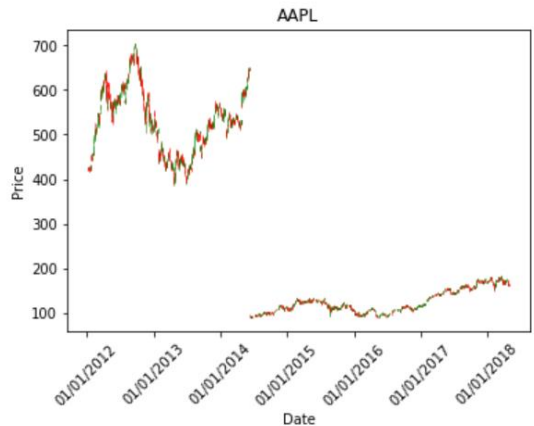

Figure 2: Apple Inc. (AAPL) split-unadjusted price movement

As follows, several steps of data preprocessing are required for cleaning the data and select the only features that model would be using for training. The downloaded historical ticker data holds following columns or features: DATE, OPEN, HIGH, LOW, CLOSE, VOLUME, EX_DIVIDEND, SPLIT_RATIO, ADJ_OPEN, ADJ_HIGH, ADJ_LOW, ADJ_CLOSE, ADJ_VOLUME, ADJ_FACTOR. However, for the model we will be using features only: OPEN, HIGH, LOW, CLOSE, VOLUME, ADJ_OPEN, ADJ_HIGH, ADJ_LOW, ADJ_CLOSE, ADJ_VOLUME. The remaining three columns were dropped because more than 99.9% of the data in those columns were either 0 or 1 and did not add any value in setting the pattern CNN is trying to learn.

| OPEN | HIGH | LOW | CLOSE | VOLUME | ADJ_OPEN | ADJ_HIGH | ADJ_LOW | ADJ_CLOSE | ADJ_VOLUME |
|---|---|---|---|---|---|---|---|---|---|
| 109.55 | 110.1599 | 109.13 | 109.40 | 8642514.0 | 109.55 | 110.1599 | 109.13 | 109.40 | 8642514.0 |
| 109.97 | 110.8200 | 109.34 | 110.10 | 10297133.0 | 109.97 | 110.8200 | 109.34 | 110.10 | 10297133.0 |
| 110.27 | 110.5300 | 108.60 | 109.99 | 13546110.0 | 110.27 | 110.5300 | 108.60 | 109.99 | 13546110.0 |
| 111.75 | 112.9000 | 109.59 | 110.41 | 16452412.0 | 111.75 | 112.9000 | 109.59 | 110.41 | 16452412.0 |

Figure 3: Cleaned data before normalization was applied

Data augmentation is used to create multiple images by sliding the window by number days to predict. This approach has enhanced the ability of the proposed model to find and learn more complex pattern that already has access to the daily data ranging from eight to twenty-four years. The whole dataset was discretized by using split and shuffle technique to process the train data discretely and prevent the overfitting.

Figure 4: Data-augmentation and sliding window

The simple normalization technique was used to get all the data on the same scale to avoid a knock-on effect on your ability to learn. Ensuring standardized feature values by normalizing the input data implicitly and weights all features equally in their representation.

$$z = \frac{x - \min(x)}{\max(x) - \min(x)}$$

The complete dataset used in training and testing of this paper has been loaded from an authorized market data source provider Intrinio.com, a Financial Data Platform for all current stock consists of S&P 500.

## 4. Methods

Deep-Learning (DL) [18] approaches that relate to computational algorithms using artificial neural networks with many layers allows to directly analyze raw data measurements without having to encode the measurements in task-specific representations. This work proposes to harness CNN networks to efficiently learn complex non-linear models directly from the raw data, for the prediction of the financial market trends. This paper introduces the following contributions:

First, a learning scheme based on the raw market data prices data of stocks, in contrast to financial forecasting schemes where the analysis utilized engineered features [19].

Second, a trading strategy that utilizes the probabilistic predictions of the convolutional neural network model, to determine the bullish or bearish trend of the equities.

Third, the model is trained on stock dividend/split data to see if the model has learned these complex market events.

A ConvNet arranges its neurons in three dimensions (width, height, depth), as visualized below in one of the layers.

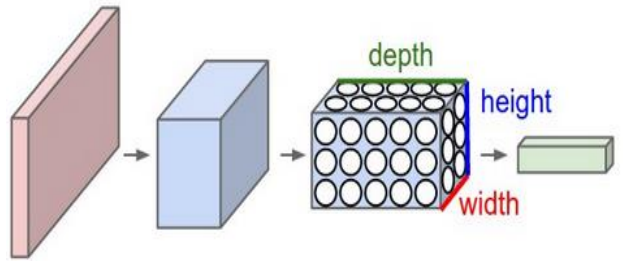

Figure 5: 1D-CNN graphical illustration [22]

In the proposed model, the convolutional layer is one dimensional, and some of the filters would be sensitive to short-term bullish trend, and they will be combined with fully connected layers to be sensitive to long-term bullish trend. For some complex patterns such as short-term bearish trends, or an overall downward trend capture CNN will handle that as well in the same way. The 10-layer model where first Conv layer has a RELU and rest of Conv layers has a Leaky RELU unit and BatchNorm layer attached is as follows:

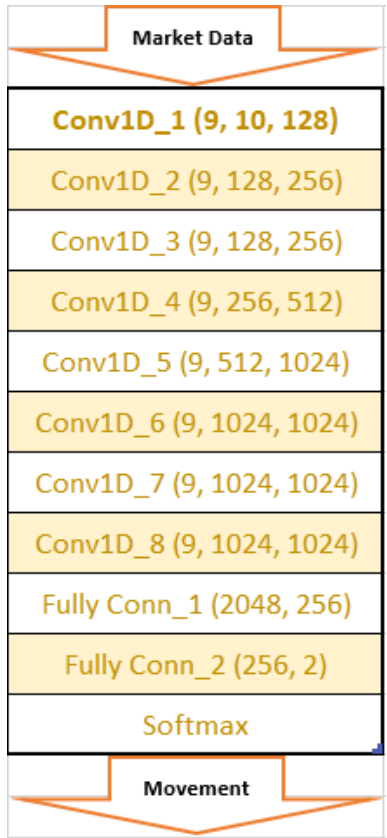

Figure 6: Cross-entropy loss graph

The input data snapshots/images are in shape [batch_size, window_size, channels], the rolling-window (the number of days of the data model is looking at in one batch) and the ten channels are OHLCV and adjusted OHLCV. After eight convolutional layers (first with RELU and rest with Leaky RELU activation) and followed by batch norms it comes to a tensor sized [batch_size, output_size, 1024], which run through several softmax layers and finally a sigmoid activation to result in a tensor sized [batch_size, output_size], where output_size is of two values, one represents the bullish confidence and the other bearish trend.

For the model proposed in this paper Softmax classifier, which has a loss function based on binary Logistic Regression, has been used. The Softmax classifier uses the cross-entropy loss where Softmax function is used that takes a vector of arbitrary real-valued scores and squashes it to a vector of values between zero and one that sum to one. Then the cross-entropy loss is applied. Cross-entropy loss, or log loss, measures the performance of a classification model whose output is a probability value between 0 and 1. Cross-entropy loss increases as the predicted probability diverge from the actual label.

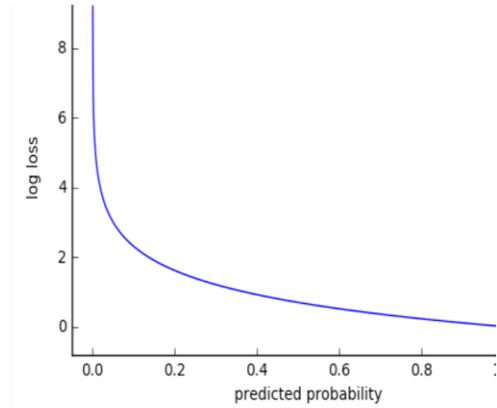

Figure 6: Cross-entropy loss graph

The graph above illustrates the range of possible loss values given an accurate observation. As the predicted probability approaches 1, log loss slowly decreases. As the expected probability decreases, however, the log loss increases rapidly. Log loss penalizes both types of errors, but especially those predictions that are confidently wrong. [21]

Below function illustrates a multiclass single fully-connected layer neural network with Softmax function:

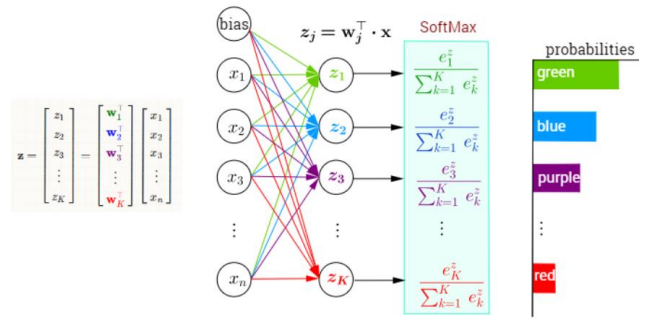

Figure 7: Classification using Softmax function (for illustrations only). Image source: stats.stackexchange.com

Softmax classifier treats the outputs f(x$_i$,W) as uncalibrated and gives a slightly more intuitive output (normalized class probabilities) and also has a

Probabilistic interpretation. In the Softmax classifier, the function mapping f(x$_i$;W)=Wx$_i$ stays unchanged, but we now interpret these scores as the unnormalized log probabilities for each class and replace the hinge loss with a cross-entropy loss that has the form:

$$L_i = -\log\left(\frac{e^{f_{y_i}}}{\sum_j e^{f_j}}\right)$$

or equivalently

$$L_i = -f_{y_i} + \log \sum_j e^{f_j}$$

where we are using the notation f$_j$ to mean the j-th element of the vector of class scores f. The full loss for the dataset is the mean of L$_i$ overall training examples together with a regularization term R(W). [22]

The *cross-entropy* between a "true" distribution p and an estimated distribution q is defined as:

$$H(p,q) = -\sum_x p(x) \log q(x)$$

The Softmax classifier is hence minimizing the cross-entropy between the estimated class probabilities and the "true" distribution, which in this interpretation is the distribution where all probability mass is on the correct class. The Softmax classifier interprets the scores inside the output vector f as the unnormalized log probabilities. Exponentiating these quantities, therefore, gives the (unnormalized) probabilities, and the division performs the normalization so that the probabilities sum to one. In the probabilistic interpretation, we are therefore minimizing the negative log likelihood of the correct class, which can be interpreted as performing *Maximum Likelihood Estimation* (MLE). [22]

The base of the model is started from an online blog by Xu, P at GitHub [23] that tries to achieve a similar goal, and except data augmentation and few parts of the model, every other piece of code has gone through a redesign and rewritten.

The proposed model implementation varies from the model referred as a base model in terms of complexity of input data. Also, the model is exposed to be trained on split/dividend unadjusted and adjusted prices, breaking input data into batches, a total number of input channels, rolling window size, CNN architecture such as adding layers and biases, dropouts, splitting the data into train-validation-test sets and shuffling it using sklearn. In addition to the aforementioned changes, the model validation set was completely missing in the base model that I have introduced here.

**5. Experiments**

BASE MODEL ACCURACIES:
The base model [23] accuracy after training the model for around an hour with different sets of data and multiple stocks the highest accuracy achieved was 69%. Additionally, to provide further perspective, following image demonstrates the prediction accuracies from some of the Deep Learning models implemented in previous papers trying to achieve a similar goal:

| Time CNN | ST-CNN 6L-3Ch | ST-CNN 6L-5Ch | ST-CNN 9L-5Ch | ST-CNN 6L-7Ch |
|---|---|---|---|---|
| 38.6 % | 41.9 % | 41.8 % | 43.9 % | 44.3 % |

| NN | RNN 1L | RNN 2L | STaR NN | STaR Linear |
|---|---|---|---|---|
| 42.1 % | 45.3 % | 44.3 % | 48.3 % | 43.2 % |

Figure 8: Base Deep models result without feature engineering [15]

INITIAL EXECUTION/TRAINING RESULTS:
After a redesign and rewriting the model following image illustrates the result of one of the initial execution result of our proposed model and it does seem to be not learning at all:

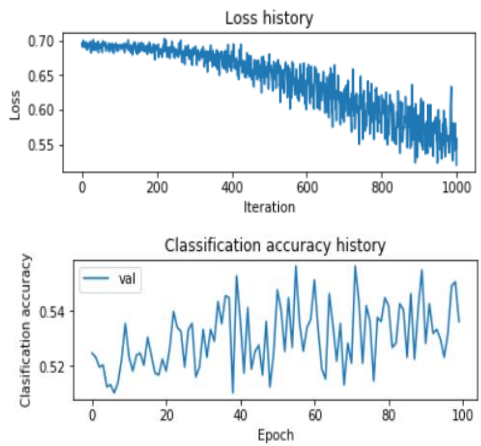

Figure 9: Initial execution loss and accuracy MSFT 2-day (T+2)

HYPERPARAMETERS TUNING:

Following are the ranges of input and hyperparameters that used are performance tuning and babysitting the model:

- Hight: 5, 64, 128, 256
- Channels: 1, 5, 10
- Learning Rates: 1e-2, 1e-3,1e-4,1e-5
- Optimizers: SGD, Adam optimizer
- Batch sizes: range 10 to 400

The optimum results were achieved when the model was trained on following input and hyperparameters:

- Channels: 10
- Hight: 256
- Width: 9
- Learning Rates: 1e-3
- Keep Prob: 0.6
- Batch size: 250
- Optimizers: Adam

Though with 256 the training took comparatively little more time than height as 128. The batch size was used ranging from 10 to 300 and while validation we observed optimum accuracy vs. performance while the batch size is ranging between 200-250. Working with a batch size between 10-150 the model execution was fast, but the learning was not good. The primary matrices used here is accuracy and loss matrices.

| Epoch | Iteration# | Training Loss | Train Acc | Validation Acc |
|---|---|---|---|---|
| 20 | 400 | 0.50728 | 0.77503 | 0.7599 |
| 40 | 800 | 0.4798 | 0.80727 | 0.8038 |
| 60 | 1200 | 0.44965 | 0.83128 | 0.8292 |
| 80 | 1600 | 0.43091 | 0.83951 | 0.8477 |
| 100 | 2000 | 0.41759 | 0.84156 | 0.8395 |

Figure 10: Loss/Accuracy table

MODEL RESULTS:

Following page contains visual graph plots of losses and accuracies generated after training the proposed method/model and results look promising where JPM (JP Morgan) stock movement forecasting accuracy touches 91% correct results indicate state-of-the-arts prediction. The proposed method outperforms the baselines previously mentioned in this paper and several models introduced by other articles published recently where the deep learning approach is being used for stock movement or price prediction.

Following are few of the graphs/plots from several experiments with the proposed model changing input, epoch, batch size and hyperparameter:

**NVIDIA** 5-day (T+5) forecasting, LR 0.0001

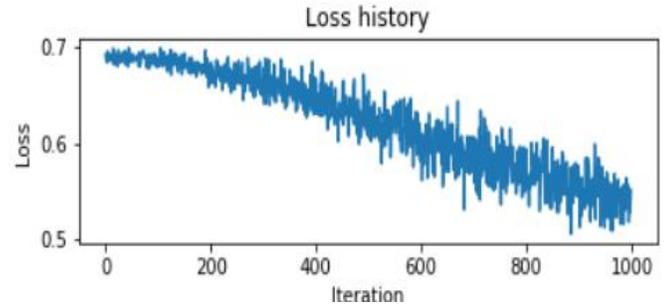

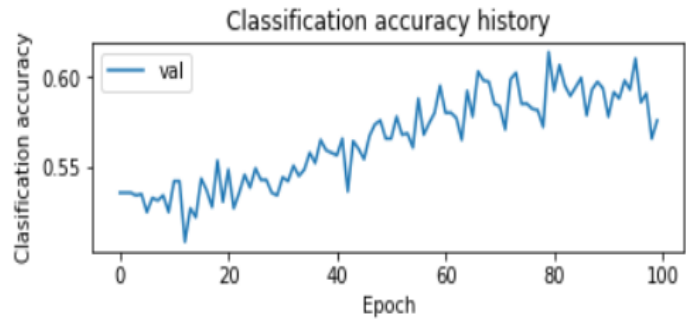

**BAC** 30-day (T+30) forecasting, LR 0.001

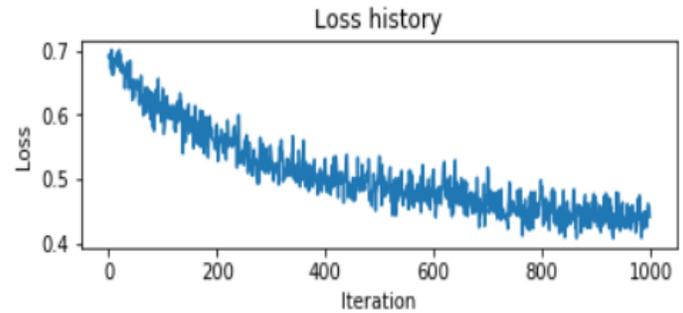

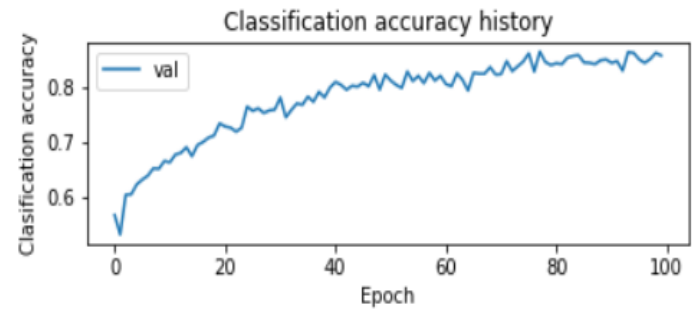

**TECH Sector** 3-day (T+30) forecasting, LR 0.001

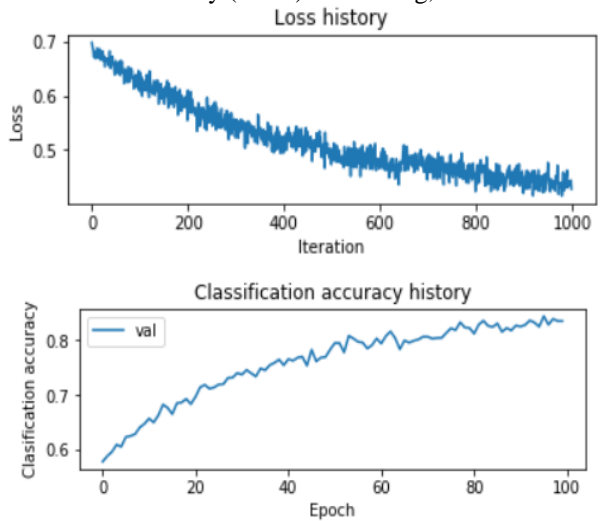

**JPM** 30-day (T+30) forcasting, LR 0.001

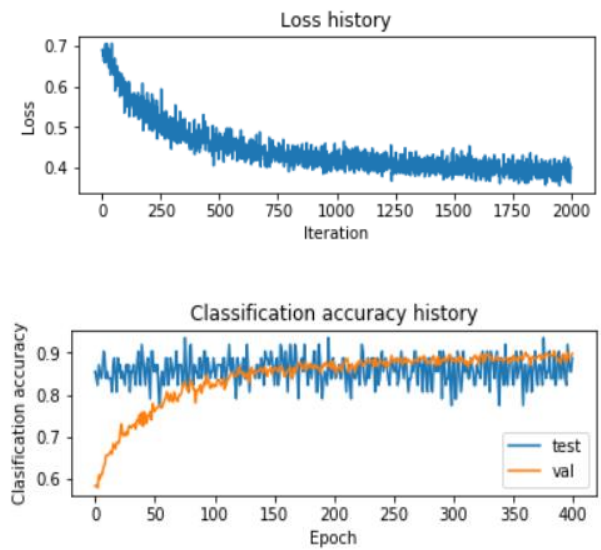

## 6. Conclusion

The results achieved by the model proposed in this paper seems promising for forecasting single stock movement as well as sector-wise progression. We can also utilize the existing model for prediction of movement of S&P 500 index once we train the model on 500 constituents historical data. There can be many ideas for future extensions or even creating new applications based on the model proposed here. It can be used for generating sets of most promising portfolios in future. The same model can be used to predict with other stocks and indices of across international capital market with minimal effort. There are numerous opportunities of future extensions such as generating sets of most profitable portfolios or predicting forex exchanges, and commodities, e.g., gas, precious metals. More sophisticated model with the CNN combined with LSTM that can manipulate its memory state and NLP to add non-financial aspects thus putting Fundamental Analysis as well in play for more accurate forecasting using sentiment analysis [20] on current affairs.

Further opportunities possible by leveraging the cutting-edge technology in a mostly unexplored by deep learning enthusiast and researchers is Derivatives Market for Options and Futures trading.